# A Weighted Multi-Criteria Decision Making Approach for Image Captioning

Hassan Maleki Galandouz, Mohsen Ebrahimi Moghaddam[*], Mehrnoush Shamsfard

Faculty of computer science and Engineering, Shahid Beheshti University G.C, Tehran, Iran
[*]m_moghadam@sbu.ac.ir

**Abstract:** Image captioning aims at automatically generating descriptions of an image in natural language. This is a challenging problem in the field of artificial intelligence that has recently received significant attention in the computer vision and natural language processing. Among the existing approaches, visual retrieval based methods have been proven to be highly effective. These approaches search for similar images, then build a caption for the query image based on the captions of the retrieved images. In this study, we present a method for visual retrieval based image captioning, in which we use a multi criteria decision making algorithm to effectively combine several criteria with proportional impact weights to retrieve the most relevant caption for the query image. The main idea of the proposed approach is to design a mechanism to retrieve more semantically relevant captions with the query image and then selecting the most appropriate caption by imitation of the human act based on a weighted multi-criteria decision making algorithm. Experiments conducted on MS COCO benchmark dataset have shown that proposed method provides much more effective results in compare to the state-of-the-art models by using criteria with proportional impact weights.

## 1. Introduction

Image captioning has recently received significant attention in the computer vision scope because it has many applications such as human-machine interaction and locating images in the form of verbal communication. Also it brings together two key areas in artificial intelligence: computer vision and natural language processing [1].

Image captioning is a much more involved task regarding to image classification, object detection, or attribute prediction, because producing a good description of an image requires a more sophisticated and holistic understanding of the image [1]. The description should consider all image visual aspects such as: objects and their attributes, scene features (e.g., indoor/outdoor), and verbalize interactions of the people and objects[2].

Image captioning methods are categorized into two main groups as follows:
Methods in the first group attempts to generate novel captions directly from images [3]–[5]. They try to recognize image content and extract information about objects, attributes, scene types, and actions, based on a set of visual features. Then, this information is used to generate the caption through surface realization. Over the last few years, a particular set of generative approaches use convolutional neural networks (CNNs) and recurrent neural networks (RNNs) to generate a caption for an image. These models first extract high-level features from a CNN trained on the image classification task, then learn a recurrent model to predict subsequent words of a caption conditioned on image features and previously predicted words [6]–[9].

The second group of approaches cast the problem to consider image captioning as a retrieval problem[2]. To make a query image description, these approaches find similar images and then build a caption for the query image using the captions of the retrieved images. The query image can be described by reusing the caption of the most similar retrieved image (transfer), or by synthesizing a novel caption using the captions of the retrieved images. Retrieval-based approaches can be categorized based on image representation images and similarity computation. In the first subgroup, image and sentence features are projected into a common multimodal space using a training set of image–description pairs. And then, the query image captions are retrieved using the multimodal space [10]–[12] while the second subgroup retrieve images using a visual space. In this group, the query image is represented by specific visual features and then a candidate set of images retrieved from the training set based on a similarity measure in the feature space. Finally, the captions of the candidate images are re-ranked to find the most appropriate caption for a query image by further use of visual and/or textual information exist in the retrieval set. , or combine fragments of the candidate descriptions according to certain rules or schemes [13]–[16]

Compared to approaches that generate captions directly, retrieval-based approaches are highly dependent on the amount of data available and the quality of the retrieval set. In order to produce image captions that are satisfactory for new test images by visual retrieval-based approaches, the quality of the retrieval set should be desirable. Also a similarity metric is needed that can measure the amount of matching between query image and retrieved captions.

It seems, the human mind, when comparing the image with the sentence, measures their similarity in terms of multiple criteria and then, assigns a weight to each of the criteria, and finally, selects the most appropriate caption based on these criteria and their weights. The main idea of our approach is also taken from this matter and in this way, we introduce a novel multi-criteria decision making step based on the impact weight for each of the criteria to improve the results. We design a mechanism to retrieve semantically more relevant captions with the query image and then select the most appropriate caption by imitation of the human act based on a multi-criteria decision making algorithm. The proposed approach considers several criteria which play a significant role in selecting the most semantically appropriate caption for the query image and selects the best caption by calculating the impact weight of the criteria. The used criteria are determining match-rate between objects, attributes and actions of the query image with nouns, adjectives, and verbs of the retrieved captions.

Experimental results of the proposed method on the MS COCO popular dataset show that our model has better results versus the related-works and produces more appropriate captions for query images.

The rest of the paper is organized as follows. We describe related-works in sec. 2, followed by a detailed description of our system in sec. 3. We report empirical results in sec. 4 and discussion and conclusion in sec. 5.

## 2. Related-works

One of the first work in retrieval-based approaches is the Im2Text model [13] which proposes a two-step retrieval process to retrieve a caption for a query image. The first step is to find visually similar images by applying some global image features. This step is baseline of most of retrieval based approaches. GIST [17] and Tiny Image[18] descriptors are employed to represent the query image. In the second step (the re-ranking step), according to the retrieved captions, some detectors (e.g., object, stuff, pedestrian, action detectors) and scene classifiers that are related to the entities mentioned in the candidate captions are employed to construct a semantic representation and re-rank the associated captions.

In the model proposed in [14], at first the authors extract and represent the semantic content of query image by applying the detectors and the classifiers used in the re-ranking step of the Im2Text model. Then, a separate image retrieval step for each detected visual element is applied on query image to collect relevant phrases from the retrieved captions. In other words, this step collects three different types of phrases. Their model extracts noun and verb phrases from captions in the training set using visual similarity among object regions detected in the training images and in the query image. Similarly, prepositional phrases are collected for each stuff detection in the query image by measuring the visual similarity of appearance and geometric arrangements between the detections in the query and training images. Also, prepositional phrases are additionally collected for each scene context detection by measuring the global scene similarity computed between the query and training images. Finally, the collected phrases for each detected object are used in integer linear programming (ILP) which considers factors such as word ordering, redundancy, etc., to generate the output caption.

The proposed method by Patterson and et. al [15] presents a large-scale scene attribute dataset for the first time in the computer vision community. They trained attribute classifiers from this dataset and showed that the responses of these attribute classifiers can be used as a global image descriptor which captures the semantic content better than the standard global image descriptors such as GIST. They proposed the baseline model by replacing the global features with automatically extracted scene attributes, and got better results in caption transfer.

Formulated caption transfer as an extractive summarization problem has been presented by Mason and Charniak [16]. This model selects the output caption by employing only the textual information in the final re-ranking step. In particular, the scene attributes descriptor of [15] are used to represent images. In this approach, at first, the visually similar images are retrieved from the training set; non-parametric density estimation are used to estimate the conditional probabilities of observing a word in query image caption. The final output caption is then obtained by using two different extractive summarization techniques that are based on the SumBasic model [19] and Kullback-Leibler divergence between the word distributions of the candidate and query captions respectively.

The authors of [20] proposed an average query expansion approach using compositional distributed semantics. To represent images, they employ features extracted from the recently proposed Visual Geometry Group convolutional neural network (VGG-CNN; [21]), trained on ImageNet. These features are the activations from the seventh hidden layer (fc7). For a query image, they first retrieve visually similar images from a large dataset of captioned images. Then, a new query based on the average of the retrieved caption distributed representations, weighted by their similarity to the input image.

The method of [22] also uses CNN activations to represent images and to determine visually similar images from the training set with the query image, carry out k-nearest neighbor retrieval. It then just like the approaches by [16] and [20] chooses a caption that best describes the images from retrieved images that are similar to the query image. Their approach differs in the way of representing the similarity between caption and choosing the best candidate in the whole set. For each retrieved caption they compute the n-gram overlap F-score between retrieved caption and each other retrieved captions. They define the caption with the highest mean n-gram overlap with the other retrieved captions as consensus caption.

## 3. The proposed approach

The proposed method is composed of the following two parts. Part one: retrieve semantically more relevant captions with the query image. Part two: choosing the most appropriate caption among the retrieved captions by imitation of the human act based on several criteria (Fig. I-1 in appendix I shows overall structure of proposed method). In addition, we describe details of these parts in the following.

### 3.1. Retrieving visually similar images

#### 3.1.1 Image representation

In visual retrieval-based approaches the quality of the initial retrieval plays a fundamental role, which makes having a good visual feature of extreme importance[20]. In this way, for representing images, we use the top-layer features of a pre-trained CNN [21], which results in a 4096-dimensional feature vector.

Our first task is to find a set of k nearest training images for each query image based on visual similarity. Therefore, one important factor for the effectiveness of the approach is having no outliers. So, instead of using a fixed neighborhood, an adaptive strategy in the similar way with [20] used to select the initial candidate set of image-caption pairs $\{(I_i, c_i)\}$. For a query image $I_q$, a ratio test is employed and only the candidates that fall within a radius defined by the distance score of the query image to the nearest training image $I_{closest}$, is considered:

$N(I_q) = \{(I_i, c_i) | \, dis(I_q, I_i) \leq (1+\varepsilon) \, dis(I_q, I_{closest}), I_{closest} = \arg\min \, dis(I_q, I_i), \, I_i \in T\}$ (1)

where N denotes the candidate set based on the adaptive neighborhood, dis represents the Euclidean distance between two feature vectors, T and $\varepsilon$ is the training set and a positive scalar value respectively.

### 3.2 Selecting semantically more relevant captions with the query image

The goal of this section is to select the more relevant captions with the query image semantically: LDA method [23], using Places-CNNs features [24], and Word2vec are some sample appropriate techniques to achieve this goal. In our experiments, we tested the proposed approach using these methods and got it that the Word2vec performs better, So Word2vec is selected for this step.

#### 3.2.1 Representing Words and Captions

In this study, the meaning of a word is represented by a vector that characterizes the context in which the word occurs in a corpus. The methods used in distributional semantics can be grouped into two: the models that are based on counting (count based models) and the models that are based on predicting (predict based models) [25]. In this paper, we use a pre-trained word2vec model [20] which is the predict-based model of [26].

Like [27], to obtain the vector representation of a caption, we first remove its stop words and then create a vector by summing up the vectors of the remaining words in the caption.

#### 3.2.2 Semantic concept detection

For a query image $Iq$, at first, visually similar images are retrieved from a large collection of captioned images. In the next step, to detect a set of semantic concepts, i.e., tags that are likely to be part of the images caption, $Iq$ is inputted to a pre-trained MIL model which predicts the words that may be nouns, adjectives, and verbs (Fig. 1). In this way, the method described by [28] is used.

Like [28], [29], in order to detect such from an image, a set of tags from the caption text in the training set is selected, then the $k$ most common words in the training captions is used to determine the vocabulary of tags. To predict semantic concepts of a given test image, this problem can be treated as a multi-label classification task.

Suppose there are N training examples, and $y_i = [y_{i1}, ..., y_{ik}] \in \{0,1\}^k$ is the label vector of the $i^{th}$ image, so if the image is annotated with tag k, $y_{ik} = 1$ otherwise $y_{ik} = 0$. Let $v_i$ and $s_i$ represent the image feature vector and the semantic feature vector for the $i^{th}$ image respectively, the cost function to be minimized is [28]:

$\frac{1}{N}\sum_{i=1}^{N}\sum_{k=1}^{M}(y_{ik} \log s_{ik} + (1 - y_{ik})\log(1 - s_{ik}))$ (2)

where $s_i = \sigma(f(v_i))$ is a K-dimensional vector with $s_i = [s_{i1}, ..., s_{ik}]$, $\sigma(\cdot)$ is the logistic sigmoid function and f($\cdot$) is implemented as a multilayer perceptron (MLP). In testing, for each input image, a semantic concept vector s is computed which is formed by the probabilities of all tags and computed by the semantic-concept detection model.

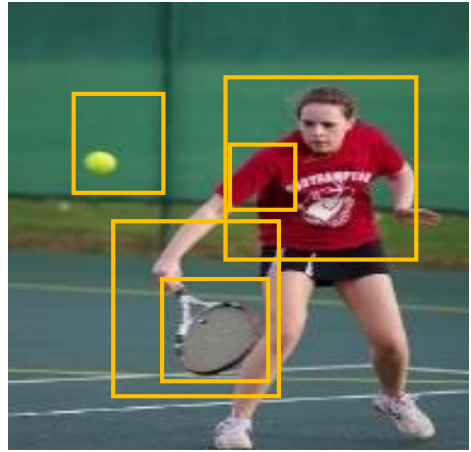

*{woman, tennis ball, red, racquet, hit}*
**Figure 1**: *A query image and predicted semantic concepts for it*

#### 3.2.2 Building MIL and q vector

After predicting image tags, their vectors are obtained by using word2vec, and then we form a general MIL vector as the sum of these vectors. This vector is a good approximation of the caption vector that should eventually be chosen for the Iq. Then, the retrieved captions are re-ranked by estimating the cosine distance between the vectors of captions and the MIL vector. Finally, n captions close to the MIL vector are selected as the candidate descriptions of the input image for more detailed investigations. The procedure to obtain the n captions close to the MIL vector is summarized in Algorithm 1.

But, there are some cases that MIL model could not predict any suitable word for query image. In these cases, instead of the MIL vector, like [13], the q vector is created based on the weighted average of the vectors of the retrieved captions as follows [20]:

$q = \frac{1}{NM}\sum_{i=1}^{N}\sum_{j=1}^{M} sim(I_q, I_i) \cdot c_i^j$ (3)

where N and M represent the total number of image-caption pairs in the candidate set N and the number of reference captions associated with each training image respectively, and $sim(I_q, I_i)$ refers to the visual similarity score of the image $I_i$ to the query image $I_q$ which is used to give more significance to the captions of images that visually are closer to the query image. $sim(I_q, I_i)$ is defined by the equation (4) as follows[20]:

$$sim(I_q, I_i) = 1 - dist(I_q, I_i)/Z^1 \quad (4)$$

Therefore, in cases where the MIL vector is constructed, n neighbors are chosen close to it; otherwise, the q vector is created and n closely neighbors to it are chosen for more detailed investigations. The number of retrieved captions that are close to the MIL vector or the q vector is varied depending on the number of images retrieved in a given radius according to equation (1).

**Algorithm1** Select semantically more relevant captions with the query image

> **Input**: Query image and retrieved captions
> **Output**: n captions close to the MIL vector
> **Begin**
> 1. Predict query image tags, T={$t_1$, $t_2$, ..., $t_m$ }.
> 2. Obtain vectors of the tags by using word2vec model, V={ $v_1$, $v_2$, ..., $v_m$ }
> 3. Compute MIL vector, MIL= $\sum_{i=1}^{m} V_i$
> 4. Compute the cosine distance between the vectors of captions and the MIL vector:
>    $D_i$= Cosine distance($C_i$ . MIL )
>    for i=1, ... ,N' (number of retrieved captions)
> 5. Sort $D_i$ descending and select top n as the n captions close to the MIL vector
>
> **End**

The maximum value of n is set to 50 because when retrieving the visually similar images to the query image, a maximum of 100 images are retrieved, and each image has 5 captions, so the total of 100 * 5 = 500 captions are retrieved. In the following, when 50 captions close to the MIL vector or the q vector are chosen, in the worst case scenario, only 50/5=10 types of descriptions may be obtained which each of them is expressed in five different forms. So, the number of neighbours is set to 50 in order to obtain at least ten different caption types.

In the next step, selected captions (candidate captions) are compared with query image in more detail based on predefined criteria including objects matching, attributes matching, and actions matching. As the result, the most appropriate caption is selected based on the above mentioned criteria and also by using a multi criteria decision making algorithm. In this way, a prepared list (prepared list_1 in Fig. 2) is used to determine the possible POS[2] tags of the MIL outputs for matching. This list is made by using the captions of the MS COCO [30] training dataset, containing 414K captions and consists of 1000 words with the highest frequency, along with the POS tag of each word in the considered caption. For words with different POS tags, their most frequent is considered.

### 3.3. How to perform matches

#### 3.3.1 Objects and actions matching

The matching of the objects in the query image with the objects (nouns) in the candidate captions is done as follows:

$$matching = \frac{M*(+1) + [min(B,Q) - M]*(-1) - [|(Q-B)|*P]}{Q} \quad (5)$$

where Q and B denote the number of objects in query image and in the candidate captions respectively and M is the number of matching between query image objects and candidate captions objects, and P refers to the amount of penalty for non-matching which calculated as follows:

$$\begin{cases} P = \frac{1}{2} & if\ Q > B \\ P = \frac{1}{3} & if\ Q < B \end{cases} \quad (6)$$

Suppose we want to get the matching rate of the query image objects with the objects of a candidate caption. According to equation (5), first the number of objects matching is obtained (M). For matches, the score +1 (M*(+1)), and for non-matching, the score -1 ([min (B,Q)-M]*(-1)) are considered. In the following, the number of query image objects and a candidate caption objects have to be checked, and if their numbers were not equal, a penalty would be imposed for it: ([| (Q-B) | * P]). But how to measure the match rate of the objects detected in the query image with the objects of candidate caption?

Consider the "motorcycle" and "bicycle" objects. These objects are not similar in appearance, but they are semantically similar. Therefore, we should use a method that can measure the semantic similarity of two words (and not the apparent similarity). Two methods to do this can be noted:

a) Use WordNet to measure the similarity of two words
b) Use the word2vec vectors of two words and measure the cosine similarity between them.

Here the second method is used. So the object matching score is obtained by measuring the cosine similarity between the word2vec vectors and equation (5). If the cosine similarity of two vectors is greater than or equal to H[3], then a match with the score of similarity has occurred. Otherwise, the non-match has happened.

The matching of actions similar to the objects matching is obtained according to the equation (5) and issues discussed above.

---

[1] Z is a normalizing constant

[2] Part of speech (POS); The NLTK POS tagger is used, which is publicly available.
[3] The parameter H is a positive real value.

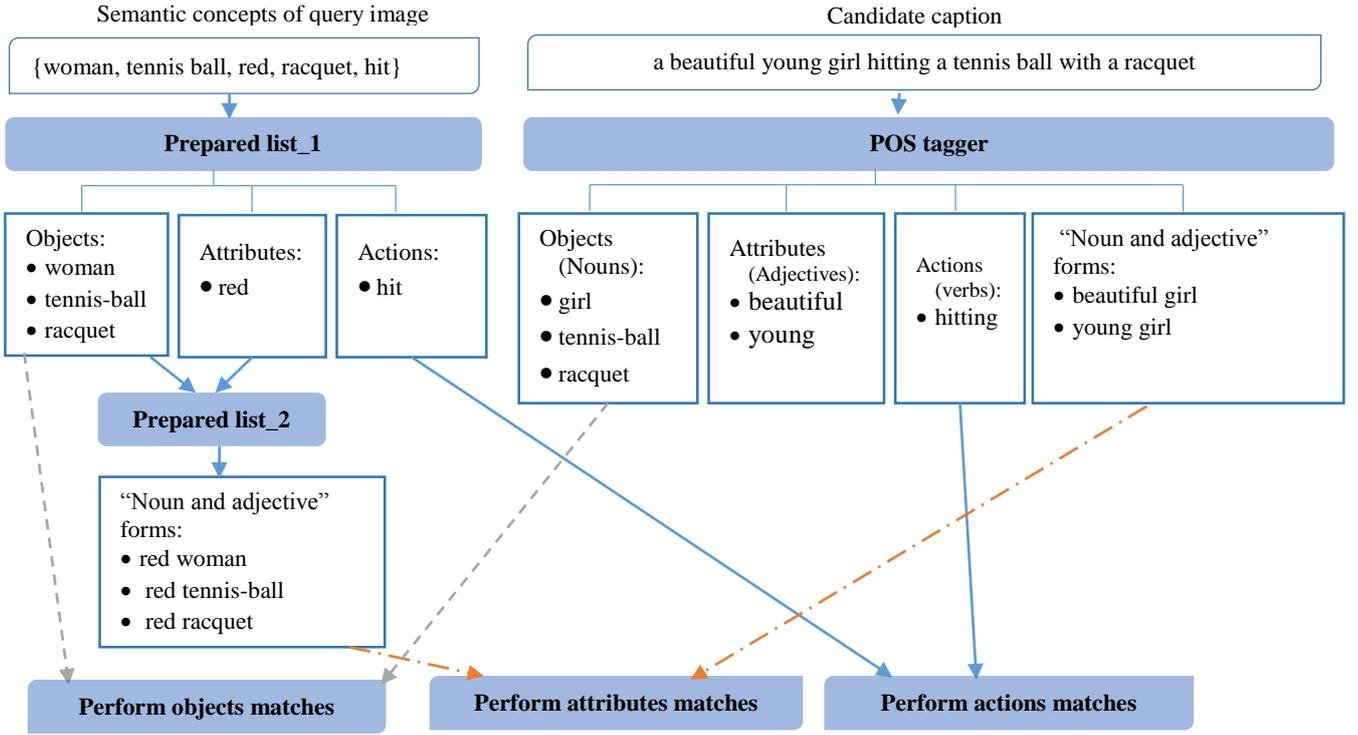

**Figure 2**: Diagram of how to perform matches

*3.3.2 Attribute matching*

Given that each attribute is related to a noun, therefore, before matching, each word derived from the MIL output which identified as an attribute is assigned to a noun. This is done through a pre-prepared list (pre-prepared list_2 in Fig. 2). This list is also made by using the captions of the MS COCO training dataset, and includes all of "noun and adjective" forms that are seen in the MS COCO training set.

On the other hand, in the candidate captions, the "noun and adjective" forms are obtained by using POS tagger tool. Finally, the "noun(object) and adjective(attribute)" forms of the query image with the "noun and adjective" forms of the candidate captions are matched by using word2vec vectors. The vector of the "noun and adjective" form is created by summing up the vectors of the constituent words. Similar objects and actions, in the case of attributes, those whose cosine similarity is equal or greater than H, are considered. Only in cases that the complete match occurs (that is, the cosine similarity is one), score one is considered, and in the other cases the similarity score is considered.

A diagram of how to perform matches is presented in Fig. 2. At this step, we should decide, based on the criteria discussed above, which of the candidate captions is the most appropriate caption for the query image.

*3.4. Multi-criteria decision making*

In most cases, decisions are desirable when decisions are made based on the several criteria. Multiple criteria decision making (MCDM) refers to making decisions in the presence of multiple and usually conflicting criteria [31]. In general, there exist two distinctive types of MCDM problems due to the different problems settings: one type has a finite number of alternative solutions and is referred to as multiple criteria decision and the other has an infinite number of solutions and is referred to as multiple objective optimization [31]. In this study, we have a multiple criteria decision problem.

A MCDM problem may be described using a decision matrix (Fig.3). Suppose there are m alternatives (captions) to be assessed based on n criteria, a decision matrix is a m × n matrix with each element $Y_{ij}$ being the $j^{th}$ criteria value of the $i^{th}$ alternative. Elements of decision matrix are filled according to the equations (5) and (6). At this point, we are ready to make decisions using the decision matrix, but the determination of the impact weights of evaluation criteria is one important step that should be considered.

| Criteria / Alternatives | $X_1$ | $X_2$ | $X_3$ | ... | $X_n$ |
|---|---|---|---|---|---|
| $A_1$ | $r_{11}$ | $r_{12}$ | $r_{13}$ | ... | $r_{1n}$ |
| $A_2$ | $r_{21}$ | $r_{22}$ | $r_{23}$ | ... | $r_{2n}$ |
| $A_3$ | $r_{31}$ | $r_{32}$ | $r_{33}$ | ... | $r_{3n}$ |
| ... | ... | ... | ... | ... | ... |
| $A_m$ | $r_{m1}$ | $r_{m2}$ | $r_{m3}$ | ... | $r_{mn}$ |

**Figure 3**: Decision matrix example

Table1. Quantitative results. In all columns, the higher numbers indicate a better performance.

|       | B-1  | B-2  | B-3  | B-4  | METEOR | ROUGE |
|-------|------|------|------|------|--------|-------|
| **OUR**     | 50.0 | 30.1 | 18.3 | 11.4 | 17.6   | 37.3  |
| *OUR_QE    | 44.7 | 24.9 | 13.9 | 7.9  | 14.2   | 32.9  |
| QE(2015)    | -    | -    | -    | 5.36 | 13.17  | -     |
| MC-KL(2014) | -    | -    | -    | 4.04 | 12.56  | -     |
| MC-SB(2014) | -    | -    | -    | 5.02 | 11.78  | -     |
| VC(2011)    | -    | -    | -    | 3.71 | 10.07  | -     |

### 3.4.1 Determine the impact weight of the criteria

In this step, the impact weight of the criteria is determined by using Shannon's entropy algorithm. Shannon's entropy is a well-known method in obtaining the weights for an MADM problem especially when obtaining a suitable weight based on the preferences and DM [3] experiments are not possible [32]. The concept of Shannon's entropy has an important role in information theory and is used to refer to a general measure of uncertainty[32]. The original procedure of Shannon's entropy can be expressed in a series of steps:

1) Normalize the decision matrix.

$$P_{ij} = \frac{r_{ij}}{\sum_{i=1}^{m} r_{ij}} \forall_{ij} \text{ for } i = 1, \ldots, m \text{ and } j=1, \ldots, n. \quad (7)$$

The raw data are normalized to eliminate anomalies with different measurement units and scales. This process transforms different scales and units among various criteria into common measurable units to allow for comparisons of different criteria.

2) Compute entropy

$$E_j = -\frac{1}{\ln m} \sum_{i=1}^{m} P_{ij} Ln(P_{ij}), \text{ For } j=1, \ldots, n \quad (8)$$

3) Set the degree of diversification as:

$$d_j = 1 - E_j, \text{ For } j=1, \ldots, n \quad (9)$$

4) Set the importance degree of attribute j:

$$W_j = \frac{d_j}{\sum_{s=1}^{n} d_s}, \text{ For } j=1, \ldots, n \quad (10)$$

In the next step, Multi-criteria decision making is done by using TOPSIS (technique for order preference by similarity to an ideal solution) algorithm.

### 3.4.2 Decision making by using TOPSIS Algorithm

Among the various methods of decision making with multiple criteria, the TOPSIS method was chosen for this study because of the advantages that it has over other methods such as the possibility of applying quantitative and qualitative criteria simultaneously. TOPSIS method is presented in [33], with reference to [34]. TOPSIS is a multiple criteria method to identify solutions from a finite set of alternatives. The basic principle is that the chosen alternative should have the shortest distance from the ideal solution and the farthest distance from the negative-ideal solution.

The procedure of TOPSIS consists of the following steps:

1) Calculate the normalized decision matrix. The normalized value $n_{ij}$ is calculated as

$$n_{ij} = \frac{r_{ij}}{\sqrt{\sum_{i=1}^{m} r_{ij}^2}} \text{ for } i = 1, \ldots, m \text{ and } j=1, \ldots, n. \quad (11)$$

2) Calculate the weighted normalized decision matrix. The weighted normalized value $v_{ij}$ is calculated as

$$v_{ij} = w_j n_{ij} \text{ for } i = 1, \ldots, m \text{ and } j=1, \ldots, n \quad (12)$$

where $w_j$ is the weight of the j$^{th}$ criterion, and $\sum_{j=1}^{n} w_j = 1$.

These weights are obtained using the Shannon's Entropy Algorithm

3) Determine the positive-ideal and negative-ideal solution

$$A^+ = \{(v_1^+, v_2^+, \ldots, v_n^+)\} = \{(\max v_{ij}|i \in O), (\min v_{ij}|i \in I)\} \quad (13)$$

$$A^- = \{(v_1^-, v_2^-, \ldots, v_n^-)\} = \{(\min v_{ij}|i \in O), (\max v_{ij}|i \in I)\} \quad (14)$$

where O is associated with benefit criteria, and I is associated with cost criteria.

4) Calculate the separation measures, using the n-dimensional Euclidean distance. The separation of each alternative from the ideal solution is given as

$$d_i^+ = \{\sum_{i=1}^{m}(V_{ij} - V_i^+)\}^{\frac{1}{2}} \forall i. \quad (15)$$

Similarly, the separation from the negative-ideal solution is given as

$$d_i^- = \{\sum_{j=1}^{m}(V_{ij} - V_i^-)\}^{\frac{1}{2}} \forall i. \quad (16)$$

5) Calculate the relative closeness to the ideal solution. The relative closeness of the alternative $A_i$ with respect to $A^+$ is defined as

---

[3] Decision Maker (DM)

* The OUR_QE method is the same as the QE method re-implemented by the authors

$cl_i = \frac{d_i^-}{d_i^+ + d_i^-}$ for i = 1, ..., m. Since $d_i^- \geq 0$ and $d_i^+ \geq 0$, then clearly $cl_i \in [0, 1]$.   (17)

6) Rank the preference order. ranking alternatives using this index in decreasing order.

## 4. Experimental setup and evaluation

Details about experimental setup, are given below.

### 4.1. Corpus

Representation of words is based on the captions of the MS COCO dataset, containing 620K captions. In the pre-processing step, all captions in the corpus are converted to lower case, and punctuation are removed. Like [20], vectors are 500-dimensional and are trained using the word2vec [26] model.

### 4.2. Dataset and Settings.

We perform experiments on the popular large scale MS COCO [30] dataset, containing 123K images. It contains 82,783 training images and 40,504 validation images. Most images contain multiple objects and significant contextual information, and each image accompanies with 5 reference captions annotated by different people. The images create a challenging testbed for image captioning and are widely used in recent automatic image captioning work. In order to compare the proposed method with previous works, we used the train, validation, and test splits prepared by [6], that is, all 82,783 images from the training set for training, and 5,000 images for validation and 5000 images for testing.

For our experiments, we utilized the corresponding validation split as a "tuning" set for hyper-parameter optimization of proposed method, and used the test split for evaluation and reporting results where we considered all the image-caption pairs in the training and the validation splits as the knowledge base. The parameter H (in section 3.3) is set to 0.85, which is obtained empirically.

MS COCO dataset is under active development and might be subject to change. In this study, results reported with version 1.0 of MS COCO dataset. We also follow the publicly available code [6] to preprocess the captions, yielding vocabulary sizes of 8791 for COCO.

### 4.3. Metrics

The proposed approach is compared with the adapted baseline model (VC) of im2text [13] which corresponds to using the caption of the nearest visually similar image, and the word frequency-based approaches of [16](MC-SB and MCKL), and model presented by [20] (QE) which use an average query expansion approach, based on compositional distributed semantics .

For a fair comparison with the above mentioned models, the same similarity metric is used, as well as the training splits for retrieving visually similar images for all models. The quality of generated captions is measured with a range of metrics, which are fully discussed in [35-36]. These metrics are: BLEU [37], METEOR [38] and ROUGE-L[36]. Each of these methods measure the agreements between the ground-truth captions and the outputs of automatic systems. We use the public python evaluation API released by the MSCOCO evaluation server.

### 4.4. Quantitative evaluation results

Quantitative results based on evaluation metrics are presented in Table 1. According to this table, the proposed approach has better outcomes than the VC, MC-SB, MC-KL and QE models.

### 4.5. Qualitative evaluation results

Fig. I-2 (in appendix I) presents some example results obtained with the proposed method on the benchmark dataset MS COCO. For a better comparison, ground truth human descriptions and a match graph of the retrieved caption with 5 reference captions of query image are provided. According to this Figure, the proposed method, using the multi-criteria decision-making mechanism, has been able to select better caption compared to other methods.

In Fig. I-3 (in appendix I), there are some cases where the proposed approach falls short. In some of those cases, although the system does not produce the most desirable results, it often is able to produce results as it could capture some of the semantic relations correctly.

In some cases, the error in the MIL outputs affects the selection of the final caption. For example, in Fig. I-3 -a, one of the words that MIL model had predicted for this image, is female with a probability of 0.18. The prediction of this word has led to search at later steps as an attribute in the retrieved captions, and the caption that has "female" to be selected as a final caption, and the value of BLEU-4 will be zero. As another example, Fig. I-3 -b shows the results of predicting the word "sandwich" with a probability of 0.16 by the MIL model, which in the next steps will cause the selection of the wrong caption.

In some other cases, using all outputs of MIL, instead of using the words that are the main aim of the image, has led to the selection of inappropriate caption as the final caption. For example, in Fig. I-3-d, the words "paper" and "person" with a probability of 0.9 and 0.97 respectively, are predicted by MIL model, and used in the next steps. But according to the image, although "Paper" and "Person" are somehow in the image, but they are not the main things in the image.

In some other cases, the weakness of the Word2vec model in making good vectors for words so that the difference between two words can be distinguished through their vectors, has led to inappropriate caption as the final caption to be selected. For example, in the Fig. I-3-c, MIL predicts two "red" and "black" words with probabilities of 0.30 and 0.25, respectively. But in the retrieved caption ("a man in a blue jacket on a snow skis") blue is mentioned ("blue jacket"). This is because the cosine similarity of the two "red" and "blue" words in the word2vec trained on the MS COCO dataset is 0.73 and the cosine similarity of the two terms "blue jacket" (extracted from the candidate caption) with the "red jacket" (obtained from the MIL output) 0 .88 is obtained. This number is greater than the threshold value (i.e. 0.85), so the matching of these two terms is accepted at 0.88, And finally the wrong caption is selected for the image.

## 5. Discussion and conclusion

One limitation in this work is the wrong prediction or the lack of word prediction of MIL model. As shown in Fig. I-1-a and Fig. I-1-b, the proposed method consists of two parts that in both parts, the MIL outputs play a significant role. MIL may only detect some of the objects, attributes, and actions in the query image or it may only identify a few objects, attributes, actions, or it cannot identify any words at all.

The words that MIL predicts are used twice: 1) In the first part of the proposed method, the MIL vector is made by MIL outputs then according to this vector the more relevant captions are selected. 2) In the second part, the MIL output is also used to check the amount of match rate between candidate captions and the query image. Therefore, the error in the MIL outputs affects the performance of both parts. Also, it may be better that the MIL output words are checked before they are used in the next steps, in terms of how much they are related to the image, so descriptions that are closer to the main aim of the image, are retrieved.

Another limitation of this work is word2vec model, which does not make good vectors for some words, so the difference between two words cannot be understood by their vectors.

Therefore, one of our future plan is improving the word2vec and the MIL model in relation to the issues recently mentioned, that can lead to further improvement of the proposed method. Our another future plan is increasing the number of criteria in the decision-making process, which can be obtained by performing specific analyses on the query image and retrieved captions.

So as a conclusion, we have presented a framework for visual retrieval based image captioning, in which we use a multi criteria decision making algorithm to effectively combine several criteria with proportional impact weights for retrieve the most relevant caption for a query image. Experiments conducted on MS COCO benchmark dataset have shown that our framework provides much more effective results compared to the other approaches by using criteria with proportional impact weights.

## 6. References

[1] X. Li, X. Song, L. Herranz, Y. Zhu and et al., "Image captioning with both object and scene information," in *Proceedings of the 2016 ACM on Multimedia Conference*, 2016, pp. 1107–1110.

[2] R. Bernardi, R. Cakici, D. Elliott and et al., "Automatic Description Generation from Images: A Survey of Models, Datasets, and Evaluation Measures.," *J. Artif. Intell. Res.(JAIR)*, vol. 55, pp. 409–442, 2016.

[3] A. Farhadi, M. Hejrati, M. A. Sadeghi and et al., "Every picture tells a story: Generating sentences from images," in *European conference on computer vision*, 2010, pp. 15–29.

[4] G. Kulkarni, V. Premraj, V. Ordonez and et al., "Babytalk: Understanding and generating simple image descriptions," *IEEE Trans. Pattern Anal. Mach. Intell.*, vol. 35, no. 12, pp. 2891–2903, 2013.

[5] M. Mitchell, X. Han, J. Dodge and et al., "Midge: Generating image descriptions from computer vision detections," in *Proceedings of the 13th Conference of the European Chapter of the Association for Computational Linguistics*, 2012, pp. 747–756.

[6] A. Karpathy and L. Fei-Fei, "Deep visual-semantic alignments for generating image descriptions," in *Proceedings of the IEEE Conference on Computer Vision and Pattern Recognition*, 2015, pp. 3128–3137.

[7] K. Xu, J. Ba, R. Kiros and et al., "Show, attend and tell: Neural image caption generation with visual attention," in *International Conference on Machine Learning*, 2015, pp. 2048–2057.

[8] X. Chen and C. Lawrence Zitnick, "Mind's eye: A recurrent visual representation for image caption generation," in *Proceedings of the IEEE conference on computer vision and pattern recognition*, 2015, pp. 2422–2431.

[9] O. Vinyals, A. Toshev, S. Bengio and et al. "Show and tell: A neural image caption generator," in *Computer Vision and Pattern Recognition (CVPR), 2015 IEEE Conference on*, 2015, pp. 3156–3164.

[10] R. Socher, A. Karpathy, Q. V Le and et al., "Grounded compositional semantics for finding and describing images with sentences," *Trans. Assoc. Comput. Linguist.*, vol. 2, pp. 207–218, 2014.

[11] A. Karpathy, A. Joulin, and L. F. Fei-Fei, "Deep fragment embeddings for bidirectional image sentence mapping," in *Advances in neural information processing systems*, 2014, pp. 1889–1897.

[12] M. Hodosh, P. Young, and J. Hockenmaier, "Framing image description as a ranking task: Data, models and evaluation metrics," *J. Artif. Intell. Res.*, vol. 47, pp. 853–899, 2013.

[13] V. Ordonez, G. Kulkarni, and T. L. Berg, "Im2text: Describing images using 1 million captioned photographs," in *Advances in Neural Information Processing Systems*, 2011, pp. 1143–1151.

[14] P. Kuznetsova, V. Ordonez, A. C. Berg and et al., "Collective generation of natural image descriptions," in *Proceedings of the 50th Annual Meeting of the Association for Computational Linguistics: Long Papers-Volume 1*, 2012, pp. 359–368.

[15] G. Patterson, C. Xu, H. Su, and J. Hays, "The sun attribute database: Beyond categories for deeper scene understanding," *Int. J. Comput. Vis.*, vol. 108, no. 1–2, pp. 59–81, 2014.

[16] R. Mason and E. Charniak, "Nonparametric Method for Data-driven Image Captioning.," in *ACL (2)*, 2014, pp. 592–598.

[17] A. Oliva and A. Torralba, "Modeling the shape of the scene: A holistic representation of the spatial envelope," *Int. J. Comput. Vis.*, vol. 42, no. 3, pp. 145–175, 2001.

[18] A. Torralba, R. Fergus, and W. T. Freeman, "80 million tiny images: A large data set for nonparametric object and scene recognition," *IEEE Trans. Pattern Anal. Mach. Intell.*, vol. 30, no. 11, pp. 1958–1970, 2008.

[19] A. Nenkova and L. Vanderwende, "The impact of frequency on summarization," *Microsoft Res. Redmond, Washington, Tech. Rep. MSR-TR-2005*, vol. 101, 2005.

[20] S. Yagcioglu, E. Erdem, A. Erdem and et al., "A Distributed Representation Based Query Expansion Approach for Image Captioning.," in *ACL (2)*, 2015, pp. 106–111.

[21] K. Chatfield, K. Simonyan, A. Vedaldi and et al., "Return of the devil in the details: Delving deep into convolutional nets," *arXiv Prepr. arXiv1405.3531*, 2014.

[22] J. Devlin, H. Cheng, H. Fang and et al., "Language models for image captioning: The quirks and what works," *arXiv Prepr. arXiv1505.01809*, 2015.

[23] D. M. Blei, A. Y. Ng, and M. I. Jordan, "Latent dirichlet allocation," *J. Mach. Learn. Res.*, vol. 3, no. Jan, pp. 993–1022, 2003.

[24] B. Zhou, A. Lapedriza, A. Khosla and et al., "Places: A 10 million image database for scene recognition," *IEEE Trans. Pattern Anal. Mach. Intell.*, 2017.

[25] M. Baroni, G. Dinu, and G. Kruszewski, "Don't count, predict! A systematic comparison of context-counting vs. context-predicting semantic vectors.," in *ACL (1)*, 2014, pp. 238–247.

[26] T. Mikolov, I. Sutskever, K. Chen and et al., "Distributed representations of words and phrases and their compositionality," in *Advances in neural information processing systems*, 2013, pp. 3111–3119.

[27] W. Blacoe and M. Lapata, "A comparison of vector-based representations for semantic composition," in *Proceedings of the 2012 joint conference on empirical methods in natural language processing and computational natural language learning*, 2012, pp. 546–556.

[28] H. Fang, S. Gupta, F. Iandola and et al., "From captions to visual concepts and back," in *Proceedings of the IEEE conference on computer vision and pattern recognition*, 2015, pp. 1473–1482.

[29] Z. Gan, C. Gan, X. He and et al., "Semantic compositional networks for visual captioning," *arXiv Prepr. arXiv1611.08002*, 2016.

[30] T.-Y. Lin, M. Maire, S. Belongie and et al., "Microsoft coco: Common objects in context," in *European conference on computer vision*, 2014, pp. 740–755.

[31] L. Xu and J.-B. Yang, *Introduction to multi-criteria decision making and the evidential reasoning approach*. Manchester School of Management Manchester, 2001.

[32] C. E. Shannon, "A mathematical theory of communication," *ACM SIGMOBILE Mob. Comput. Commun. Rev.*, vol. 5, no. 1, pp. 3–55, 2001.

[33] D. L. Olson, "Comparison of weights in TOPSIS models. Mathematical and Computer Modeling, 40 (7-8), 721–727." 2004.

[34] K. Yoon and C.-L. Hwang, *Multiple attribute decision making: methods and applications*. SPRINGER-VERLAG BERLIN AN, 1981.

[35] D. Elliott and F. Keller, "Comparing automatic evaluation measures for image description," in *Proceedings of the 52nd Annual Meeting of the Association for Computational Linguistics: Short Papers*, 2014, vol. 452, no. 457, p. 457.

[36] R. Vedantam, C. Lawrence Zitnick, and D. Parikh, "Cider: Consensus-based image description evaluation," in *Proceedings of the IEEE conference on computer vision and pattern recognition*, 2015, pp. 4566–4575.

[37] K. Papineni, S. Roukos, T. Ward and et al. , "BLEU: a method for automatic evaluation of machine translation," in *Proceedings of the 40th annual meeting on association for computational linguistics*, 2002, pp. 311–318.

[38] M. Denkowski and A. Lavie, "Meteor universal: Language specific translation evaluation for any target language," in *Proceedings of the ninth workshop on statistical machine translation*, 2014, pp. 376–380.

## 7. Appendix I

Fig. 1 shows the conceptual diagram of our proposed approach for image captioning. Fig. 2 and Fig. 3 show some example input images and the retrieved descriptions for them, that the proposed method has produced a good and bad output compared to other methods respectively.

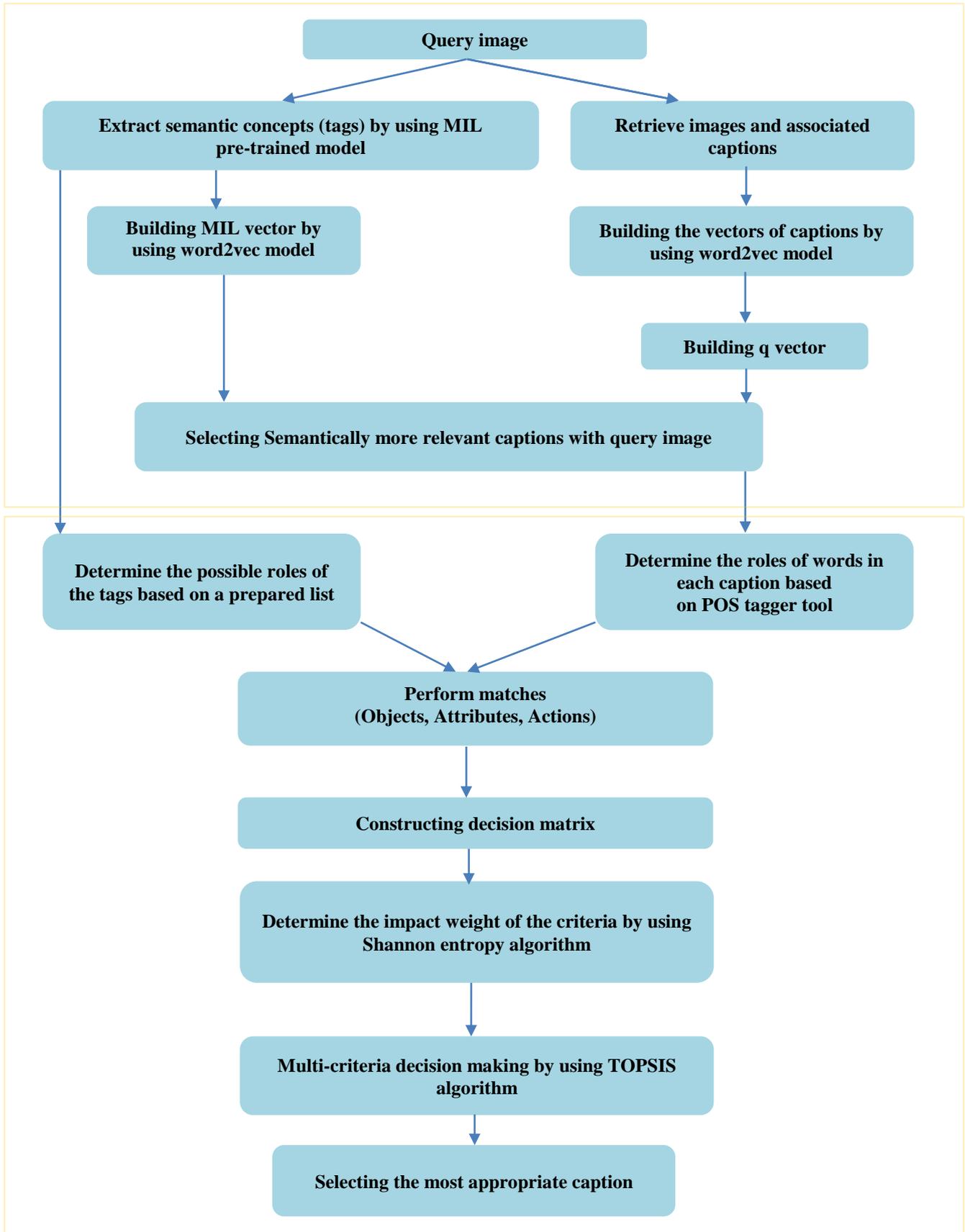

*Figure 1*: The conceptual diagram of our proposed approach for image captioning which consists of two parts; part one (a): retrieve semantically more relevant captions with the query image, part two (b): selecting the most appropriate caption among the candidate captions.

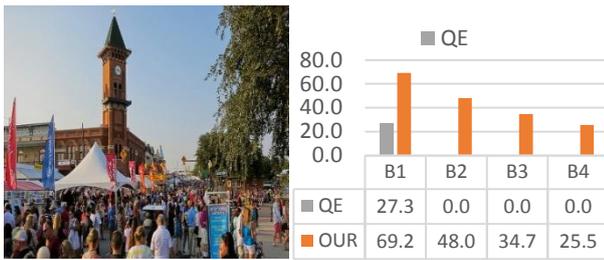
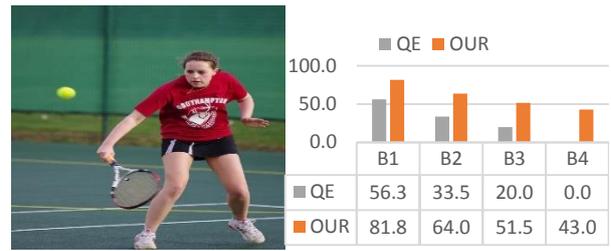

| QE | the flags of many nations flying by big ben in london |
|---|---|
| OUR | landscape of a clock tower attached to a large building in a city |
| HUMAN | a large crowd is attending a community fair |
| | a crowd of people walking in an outdoor fair |
| | a crowd of people at a festival type event in front of a clock tower |
| | the building has a clock displayed on the front of it |
| | a festival with people and tents outside a clock tower |

| QE | a small boy holding a tennis racket intently stares at a tennis ball in the air |
|---|---|
| OUR | a beautiful young woman hitting a tennis ball with a racquet |
| HUMAN | a woman hitting a tennis ball on a court |
| | a woman swinging a tennis racquet towards a tennis ball |
| | a female tennis player finishes her swing after hitting the ball |
| | a woman bending slightly to hit a tennis all with a racket |
| | a female in a red shirt is playing tennis |

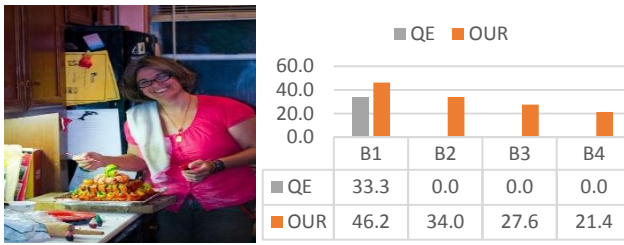
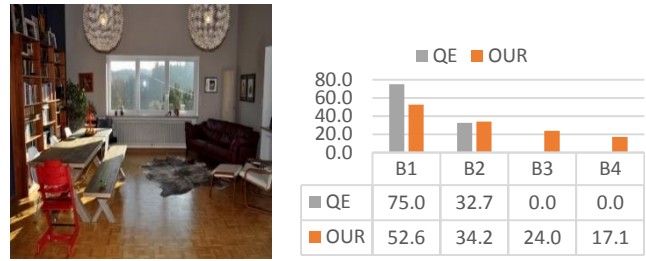

| QE | a man and boy blow out a candle on a birthday cake |
|---|---|
| OUR | an older woman sits in front of a cake near a young woman |
| HUMAN | a woman standing over a pan filled with food in a kitchen |
| | a woman smiling while she prepares a plate of food |
| | a smiling woman standing next to a plate of food she made |
| | a woman in a bright pink summer shirt smiles and displays a party platter she has made |
| | a person standing in front of a counter top and a tall pile of food |

| QE | a chair and a table in a room |
|---|---|
| OUR | modern living room with a ceiling fan two couches a coffee table a fireplace and a large screen tv |
| HUMAN | a little room and dining room area with furniture |
| | a living room with a big table next to a book shelf |
| | a living room decorated with a modern theme |
| | a living room with wooden floors and furniture |
| | the large room has a wooden table with chairs and a couch |

**Figure 2**: Some example input images and the retrieved descriptions for them, that the proposed method has produced a good output compared to other methods

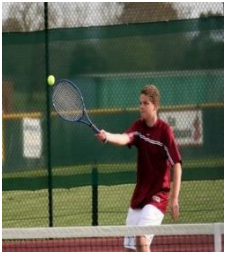 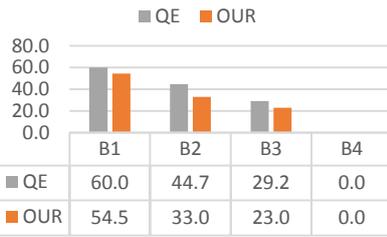 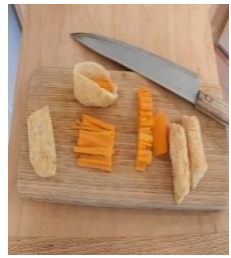 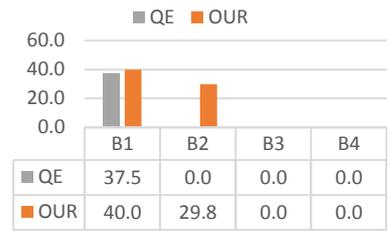

| QE | a tennis player swings his racket at a tennis ball |
|---|---|
| OUR | a female tennis player lunges forward to return the tennis ball |
| HUMAN | a guy in a maroon shirt is holding a tennis racket out to hit a tennis ball |
| | a man on a tennis court that has a racquet |
| | a boy hitting a tennis ball on the tennis court |
| | a person hitting a tennis ball with a tennis racket |
| | a boy attempts to hit the tennis ball with the racquet |

| QE | many different types of vegetables on wooden table |
|---|---|
| OUR | a sub sandwich on a wooden tray on a table |
| HUMAN | a wooden cutting board with cheese bread and a knife on it |
| | a cutting board topped with cheese bread and a knife |
| | a cutting board with carrots and thin breading |
| | sliced bread and cheese sits on a cutting board with a sharp knife |
| | carrots bread and knife on top of cutting board |

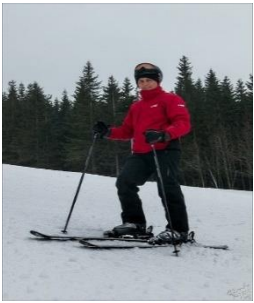 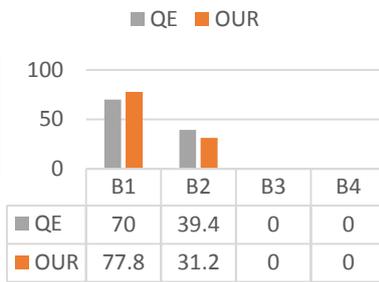 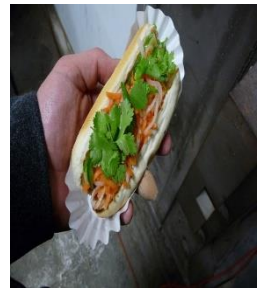 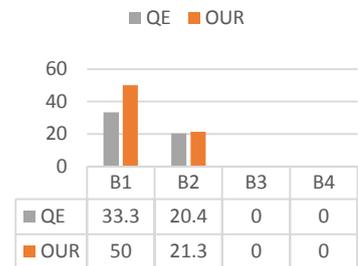

| QE | a man wearing skis at the bottom of a slope |
|---|---|
| OUR | a man in a blue jacket on snow skis |
| HUMAN | a man on skis is posing on a ski slope |
| | a person on a ski mountain posing for the camera |
| | a man in a red coat stands on the snow on skis |
| | a man riding skis on top of a snow covered slope |
| | a lady is in her ski gear in the snow |

| QE | a couple slices of pizza on a cardboard box |
|---|---|
| OUR | a person is holding a large paper box with food in it |
| HUMAN | hot dog on a roll with cheese onions and herbs |
| | a sandwich has cilantro carrots and other vegetables |
| | a hotdog completely loaded with onions and leaves |
| | a hand holding a hot dog on a bun in a wrapper |
| | the hotdog bun is filled with carrots and greens |

| a | b |
|---|---|
| c | d |

**Figure 3**: Some example input images and the retrieved descriptions for them, that the proposed method has produced

12